\def\year{2020}\relax
\documentclass[letterpaper]{article} 
\usepackage{aaai20}  
\usepackage{times}  
\usepackage{helvet} 
\usepackage{courier}  
\usepackage[hyphens]{url}  
\usepackage{graphicx} 
\usepackage{graphicx}  
\usepackage{booktabs}
\usepackage{amsmath}

\usepackage{appendix}

\newcommand{\ignore}[1]{}
\newcommand{\citet}[1]{\citeauthor{#1} \shortcite{#1}}
\newcommand{\citealp}[1]{\citeauthor{#1} \citeyear{#1}}

\nocopyright
\title{Consumer-Driven Explanations for Machine Learning Decisions:\\
An Empirical Study of Robustness
}

\author{Michael Hind, Dennis Wei, Yunfeng Zhang\\IBM Research}

\begin{document}

\maketitle

\begin{abstract}
Many proposed methods for explaining machine learning predictions are in fact challenging to understand for non-technical consumers. This paper builds upon an alternative consumer-driven approach called TED that asks for explanations to be provided in training data, along with target labels. Using semi-synthetic data from credit approval and employee retention applications, experiments are conducted to investigate some practical considerations with TED, including its performance with different classification algorithms, varying numbers of explanations, and variability in explanations. A new algorithm is proposed to handle the case where some training examples do not have explanations. Our results show that TED is robust to increasing numbers of explanations, noisy explanations, and large fractions of missing explanations, thus making advances toward its practical deployment.
\end{abstract}

\pdfinfo{
/Title (Type Your Paper Title Here in Mixed Case)
/Author (John Doe, Jane Doe)
/Keywords (Input your keywords in this optional area)
}

\section{Introduction}
\label{sec:intro}

As AI becomes more prevalent in society, many stakeholders are requiring AI predictions to be accompanied by explanations.  In addition to GDPR~\cite{gdpr-interpretation-2017}, the state of Illinois recently passed a law about video interviews that requires employers to provide information {\em ``explaining how the artificial intelligence works}~\cite{Illinois-2019}.  Concurrently, the research community has developed many explainability or interpretability techniques.  Most of these take an ``inside out'' (model-to-consumer) approach to providing explanations based on how the models work.  Although this can be useful to those explanation consumers who understand ML models, it can pose a challenge for non-technical consumers.  For example, many techniques provide explanations composed of input features of the model, e.g.~proprietary risk scores in a loan application scenario, but often these features may not be meaningful to an end user.  Other approaches explain by example by showing similar instances from the training set, which can be useful to a domain expert, but not to an end user.

The TED explainability framework~\cite{TED} takes a different ``outside in" (consumer-to-model) approach to explainable AI.  It attempts to provide meaningful explanations by capturing the conceptual domain of the explanation consumer.
The framework uses a training dataset that includes
explanations ($E$), in addition to features ($X$) and target labels ($Y$), to 
create a model that simultaneously produces predictions and explanations.
The framework is general in
that it can be used with many supervised machine learning algorithms, thereby leveraging their well-established predictive abilities to also predict explanations. 

Previous work has shown the efficacy of this approach 
on two synthetic datasets~\cite{TED} and on two other datasets where explanations are synthetically created~\cite{TED-HLL19}.  Based on the promising results from this initial work, this paper more deeply explores some of the practical considerations with the TED approach.

The contributions of this paper are 
\begin{itemize}
\item Empirical evaluations that address open questions regarding how TED performs with different base classification algorithms, with varying numbers of explanations, and with variability in the explanations for a given set of features. Results show that accuracies are maintained as the number of explanations increases to more than 30 and degrade gradually as variability increases.

\item A new algorithm that allows TED to be deployed when some training examples are missing explanations, addressing one of the most significant challenges of using TED. An evaluation of this partially supervised algorithm shows that it can reduce the number of training explanations by $90\%$ with little loss in accuracy.
\end{itemize}

These contributions demonstrate that TED is a robust 
explanation technique and useful when meaningful explanations are paramount.
Section~\ref{sec:TED} provides an overview of the TED framework. Section~\ref{sec:ted-partial} describes the new algorithm for partially supervised learning.  Section~\ref{sec:data} describes two datasets and the synthesis of labels that we use for our evaluation.  Section~\ref{sec:expt} describes our experiments and results.  Section~\ref{sec:discuss} discusses open problems and future work.

\section{The TED Explainability Framework}
\label{sec:TED}
The main motivation for the TED (Teaching Explainable Decisions) framework is that a meaningful explanation is quite subjective~\cite{Hind2019,AIX360-arxiv}, and thus, requires additional information from the explanation consumer regarding what is meaningful to them.
\ignore{
Let $X\times Y$ denote the input-output space, with $p(x,y)$ denoting the joint distribution over this space, where $(x,y)\in X\times Y$. In supervised learning one wants to estimate $p(y|x)$.
}
This additional information takes the form of training data composed of triples $X\times Y\times E$ corresponding to the input space, output space, and explanation space, respectively. From this training data, TED learns to predict $Y$ and $E$ jointly from $X$. 

\ignore{Thus, we not only predict the labels $y$, but also the corresponding explanations $e$ for the specific $x$ and $y$ based on training explanations given by human experts.}

There are many instantiations of the TED framework.  The initial paper~\cite{TED} describes a simple Cartesian product approach, where the $Y$ and $E$ labels are combined into a new label, $(Y,E)$, and a variety of supervised learning algorithms can be applied to solve the resulting multi-class problem. 
Other instantiations include 1) a multi-task learning approach in which predicting labels and explanations constitute two or more tasks, and 2) building upon the
tradition of similarity metrics, case-based reasoning and
content-based retrieval~\cite{TED-HLL19}.  

This work focuses on the Cartesian product instantiation of TED, referred to as TED-C, and implemented in the AI Explainability 360 open source toolkit~\cite{AIX360-arxiv}.  We choose this instantiation for its simplicity and generality as discussed above. 
We pursue the following open research questions:

\begin{enumerate}
\item How does TED-C impact target label ($Y$) accuracy compared to the base model?
\item How does the effectiveness of TED-C vary with different learning algorithms?
\item How does the effectiveness of TED-C scale with an increase in the number of possible explanations?
\item How does the effectiveness of TED-C depend on variability in the explanations associated with a given input $X$?
\item How can the TED framework be extended to work in a partially-supervised manner, i.e., when some training instances do not have an explanation?
\item What is the performance of this new algorithm?
\end{enumerate}

\ignore{
The
resulting model will produce predictions ($YE$) that can be
decomposed to the $Y$ and $E$ component.  This instantiation is
evaluated on two synthetic datasets for playing tic-tac-toe and
loan repayment.  Their results show
``{\em (1) To the extent that user explanations follow simple logic, very
high explanation accuracy can be achieved; (2) 
Accuracy in predicting $Y$ not only does not suffer but actually improves.}''
}

TED is best suited for use cases where the labels and explanations are consistent, which typically occurs when they are created at the same time. We explore this scenario in this paper.  In the situation where a dataset already has labels, either from a human or from actual outcomes, such as loan defaults, it can be challenging to produce explanations that are consistent with the labels.  For example, asking someone to explain why a person defaulted on a loan, given their loan application, is not a trivial matter.  We leave such scenarios for future work, noting however that the partially supervised approach described in Section~\ref{sec:ted-partial} can help reduce the amount of explanation labeling required.

\ignore{
1) no labels or explanations exist, SME needs to produce both based on their experience or rules.  This is the scenario the labeling company I spoke to lives in, although they are not doing explanations now because their clients are not asking for them.  In this scenario, I would expect the Ys and Es to be consistent, except for human error,  2) Y's already exist (either as real outcomes or by some prior labeling exercise) and someone needs to create the E's.  Currently, no one does this because they get no value, but TED requires it.  Here, we are asking a person to explain some other entity's decision, which will be very hard, even before TED enters the picture.  Concrete example:  "Please add explanations to the ImageNet dataset".    I'm not sure someone would pursue this.

I think we should at least describe these two scenarios clearly up front and say that TED mostly/exclusively addresses the first scenario. I think this will decrease confusion over what we mean by Y labels and why we expect E labels to be consistent with them. I also think we should say something about the second scenario in the Discussion/Open Problems section, including the point that asking someone to explain another entity's decision is difficult.}

One of the strengths of the TED approach is that it enables explanations that involve higher-level concepts that are more meaningful than features.   For example, in Table~\ref{table:retention:exp} we enumerate 8 different conceptual reasons (the non-indented descriptions in the table) that someone might be a retention risk, i.e., likely to leave a company.  These concepts are independent of the features and, in fact, can be implemented differently for different companies or the same company over time.  They are likely to be more meaningful to a non-technical explanation consumer.

\section{TED with Partial Supervision}
\label{sec:ted-partial}
One significant drawback of the TED framework is the requirement that training datasets contain explanations in addition to decision labels.  Although researchers have reported that producing
explanations at the same time as producing labels incurs negligible overhead~\cite{Zaidan07using-annotator,zaidan-eisner:2008:gen,DBLP:journals/corr/ZhangMW16,McDonnel16why-relevant}, it is desirable to reduce this overhead.

\ignore{
Another potential concern with the TED-C framework is the additional burden placed on human experts to provide explanation labels as well as decision labels. In particular, we assume that labelers are accustomed to providing decisions but may be less able to give explanations for all training examples.
We simulate this with the FICO and retention datasets by removing at random a fraction of the $E$ labels from the training sets.}

We approach this problem by allowing some training instances to not have explanations. We propose to use imputation to fill in missing $E$ values and thereby allow the general TED framework to be used without further modification. Using the part of the training data with $E$ labels, a model is trained to predict $E$ from $X$ and $Y$ (the latter is available for all training examples). This model is used to impute $E$ labels for the part of the training data that lacks them. The full training set with given and imputed $E$ labels is then passed as input to the TED framework.  We evaluate this approach for TED-C in Section~\ref{sec:expt:partial}.

\ignore{
To simplify our experiment, the same classification algorithm (e.g.~LGB, SVM) is used for imputation as for TED-C. For imputation, two separate models are trained for $Y=0$ and $Y=1$ (with $X$ as input), instead of appending $Y$ as an input alongside $X$. This takes advantage of the fact that $Y$ is binary and restricts the set of possible $E$'s to those associated with either the positive or negative class.
}

\ignore{
Outline

\noindent
- Logically partition partially E-labeled dataset
- Use fully E-Labeled part of dataset to train a model for predicting E from X,Y  ($p(e | x, y)$)\\
- Use model to impute missing E labels in non E-Labeled part of dataset, producing a dataset with labels\\
- Join E-imputed dataset with E-Labeled dataset as input to TED-Cartesian
}

\section{Datasets and Label Synthesis}
\label{sec:data}

Evaluation of TED requires training data containing explanations, ideally provided by domain experts. Since such datasets are not yet readily available, we instead use semi-synthetic data generated as described in this section. Semi-synthetic data does have advantages over real-world data in enabling the study of certain effects on the performance of TED, in particular those due to the amount of variability in explanations and the number of possible explanations, which are the subject of this paper (Sections~\ref{sec:expt:number} and \ref{sec:expt:noise}). 

We use two datasets described in the following subsections. For both datasets, we first derive a set of rules that are intended to approximate a domain expert's knowledge and the decisions and explanations that they might provide. We then retain the original features and synthesize decisions and explanations according to these rules. This rule-based approach is becoming increasingly popular due to the scarcity of true experts for labeling~\cite{snorkel-2017} and is common in behavioral science~\cite{behavorial-code-2014}.

\ignore{[anecdotal evidence for rule-based labelling by experts]}

\subsection{FICO Challenge Dataset}
\label{sec:data:FICO}

The FICO Explainable Machine Learning Challenge dataset \cite{fico-challenge-2018} contains credit information on around 10,000 applicants for home equity lines of credit (HELOC), together with an outcome label of whether they were delinquent for 90 days or longer in repaying. Our goal is to simulate the credit approval decisions and explanations that a loan officer might produce based on applicant features. 

To substitute for domain expertise, we used the Boolean Rules via Column Generation (BRCG) algorithm \cite{BRCG} as 
implemented in \cite{AIX360-website} to learn a set of rules for approving applications. We first removed the credit score--like `ExternalRiskEstimate' feature, which was causing explanations to be more opaque, and processed special values as detailed in Appendix~\ref{sec:data:FICOadd}. 
A $10$-fold cross-validation (CV) experiment showed that BRCG regularization parameter values of $\lambda_0 = \lambda_1 = 10^{-5}$ result in good accuracy in predicting outcomes with a simple rule set. 

The rule set learned by BRCG from the entire dataset with $\lambda_0 = \lambda_1 = 10^{-5}$ has the form shown below but with 
different thresholds (see Appendix~\ref{sec:data:FICOadd}). 
To make the rules more human-like, we replaced the thresholds with nearby round numbers to yield 
\begin{small}
\begin{enumerate}
    \item NetFractionRevolvingBurden $\leq 60$ AND\\ PercentTradesNeverDelq $> 85$ AND\\ AverageMInFile $> 48$ AND\\ MaxDelq2PublicRecLast12M $> 5$; OR
    \item NetFractionRevolvingBurden $\leq 40$ AND\\ MSinceMostRecentInqexcl7days $> 24$\footnote{This condition appears to be coded as MSinceMostRecentInqexcl7days $= -8$ in the original dataset.}.
\end{enumerate}
\end{small}
This rule set has an accuracy of $71\%$ in predicting the repayment outcome label. The `NetFractionRevolvingBurden' feature is the applicant's revolving (e.g.~credit card) debt as a percentage of their credit limit. We label a value no greater than $40$ as ``low debt'' and between $40$ and $60$ as ``medium debt''. Conditions 2 and 4 in rule 1 pertain to delinquencies; condition 2 requires the percentage of never-delinquent trades (credit agreements) to be above $85\%$ while condition 4 requires no delinquencies in the last 12 months. Condition 3 in rule 1 requires the average age of existing accounts to be greater than 4 years. While rule 1 may be regarded as a rule for general creditworthiness, rule 2 is a ``shortcut'' in checking for only two conditions, low debt and no credit inquiries in the last 2 years, as signals of ability to repay. 

Rules 1 and 2 above are used to synthesize decision ($Y$) and explanation ($E$) labels alongside the original features ($X$). The ``base'' version of this semi-synthetic FICO dataset has 9 possible explanations: 2 for the positive (approve) class corresponding to which of the two rules is satisfied (rule 1 takes precedence if both), and 7 for the negative (deny) class that are minimally sufficient in explaining why neither of the rules is satisfied. The latter 7 explanations are 1) ``high debt'' ($> 60$), which is sufficient to violate both rules, 2)--4) medium debt coupled with the violation of one of conditions 2--4 in rule 1 (earlier conditions take precedence if more than one is violated), and 5)--7) low debt but an inquiry in the last 2 years, again coupled with one of conditions 2--4 in rule 1.

To obtain larger numbers of explanations, we increase the maximum number of unsatisfied conditions that are communicated in a credit denial. The 7 explanations above for the denied class consist of only 1 or 2 unsatisfied conditions (e.g.~high debt, medium debt plus one other condition), even if there are additional unsatisfied conditions. The set of explanations may be enlarged, for example, by telling high-debt applicants that they also have too many delinquent trades or insufficient credit history. In this manner, the number of explanations can be increased to more than 30. 

\subsection{Employee Retention Dataset}
\label{sec:data:retention}
In the employee retention use case, a company tries to determine which of its employees are risks to leave and thus, for whom it should consider a proactive action to retain the employee.
We use the dataset described in ~\cite{retention-2012}, which is modeled based on real data.  The original dataset contains 14 features and 9,999 employees.  We selected the following subset of 8 features: 
{\em Business unit, Position, Salary, Months since hire, Months since promotion, Last evaluation, Evaluation slope over the last 3 years, Strong potential indicator}.  

We then created 26 rules, further grouped in categories as described below, that signify a possible retention issue.  The rules were created by an experienced manager with 17 years experience (one of the authors) and enhanced by the BRCG algorithm~\cite{BRCG} mentioned above.  The rules were then applied to the dataset to produce $Y$ labels specifying if the employee was a retention risk.  The explanation for the prediction was the rule number for retention risks and a default ``No Risk" explanation for non-retention risks.   This use case and the rules are inspired by the Proactive Retention tutorial in \cite{AIX360-website}. 

Before applying the rules, we applied discretization to several of the features. The 33 organizations were mapped to 3 groups of high-, medium-, and low-attrition organizations.  The positions were similarly grouped into 4 categories: Entry, Junior, Senior, Established.  Salary was mapped into 3 salary competitiveness groups (low, medium, high) based on how it compared to the average salary for the position in the dataset (medium was within 20\%).  Similarly, evaluations were grouped into low, medium, high.  The duration-related features were unchanged; the rules specify bounds, such as $120 <$ {\tt MonthsSincePromotion} $< 180$, to signify an employee who might be at risk for mid-career crisis.

\begin{table}[h]
\begin{center}
\begin{tiny}
\caption{Employee Retention Explanations}
\label{table:retention:exp}
\begin{tabular}{|l|} \hline
Promotion Lag  \\
 \hspace{1em} + High Attrition Organizations \\
   \hspace{2em} + Junior-level \\
   \hspace{2em} + Senior-level, high potential \\
 \hspace{1em} + Medium Attrition Organizations \\
   \hspace{2em} + Entry-level \\
   \hspace{2em} + Junior-level, Strong Perf., high potential \\
   \hspace{2em} + Senior-level, Strong Perf., low comp., high potential \\
 \hspace{1em} + Low Attrition Organizations \\
   \hspace{2em} + Entry-level \\
   \hspace{2em} + Junior-level, Strong Perf., high potential \\
   \hspace{2em} + Senior-level, Strong Perf., low comp., high potential \\ \hline
New Employee\\
 \hspace{1em} + High Attrition Organizations \\
    \hspace{2em} + Entry-level \\
    \hspace{2em} + Junior-level \\
    \hspace{2em} + Senior-level \\
 \hspace{1em} + Medium Attrition Organizations \\
    \hspace{2em} + Entry-level \\
    \hspace{2em} + Junior-level \\ \hline
Disappointing Evaluation\\
 \hspace{1em} + High Attrition Organizations \\
 \hspace{1em} + Medium Attrition Organizations \\ \hline
Compensation Doesn't Match Evaluation\\
 \hspace{1em} + High Attrition Organizations, High Evaluation \\
 \hspace{1em} + High Attrition Organizations, Medium Evaluation \\ \hline
Part of Company Acquisition\\
 \hspace{1em} + High Attrition Organizations \\
 \hspace{1em} + Medium Attrition Organizations \\ \hline
Company Not Right Fit\\
 \hspace{1em} + High Attrition Organizations \\
 \hspace{1em} + Medium Attrition Organizations \\
   \hspace{2em} + Junior-level \\
   \hspace{2em} + Senior-level \\
 \hspace{1em} Low Attrition Organizations \\ \hline
 Mid-Career Crisis (High Attrition Organizations)\\ \hline
Seeking Higher Salary (High Attrition Orgs, Good Perf.)\\ \hline
\end{tabular}
\end{tiny}
\end{center}
\end{table}

Table~\ref{table:retention:exp} describes the explanations used for the employee retention dataset.  The explanations are partitioned into 3 levels (indicated by indentation), depending on the level of detail required by the consumer.  For example, at the lowest level we have the explanations at the outermost indentation level: ``Promotion Lag", ``New Employee", ``Disappointing Evaluation", etc.  These show the top 8 reasons why employees are in danger of leaving the company.  The next level provides additional refinement of these explanations.  For example, ``Promotion Lag" is decomposed into 3 different types, depending on the organization.  Employees in Organization 1 historically have a higher attrition rate and those in Organization 3 have a lower rate, so it is useful to provide this extra level of information in an explanation. This results in a total of 16 explanations.  The last 2 explanations, ``Mid-Career Crisis" and ``Seeking Higher Salary", are not decomposed, so they remain as in the first level.  The third level provides further differentiation in explanations for those consumers who need even more details.

A key advantage of TED over other explainability approaches is the ability to define concepts that provide higher-level semantic meaning than features.  For example, the concept of ``Promotion Lag" refers to an employee believing that the time since their most recent promotion is excessive.  Exactly how this concept is mapped into features may vary among organizations, positions, etc.  For example, ``recent" can be defined as within the last $N$ months for values of $N$ that depend on organizations, positions, etc.  This is how the promotion lag rules are implemented for this use case.

\subsection{Adding Variability to Labels}
\label{sec:data:noise}

Explanations based on the rules in Sections~\ref{sec:data:FICO} and \ref{sec:data:retention} exhibit no variability in that they are a deterministic function of the features. This may not be representative of explanations obtained from humans, who may make errors, disagree with each other, especially in borderline cases, or generally vary in their assessments. To simulate these effects, we experiment with adding variability to the synthesized explanations, which we will refer to as \emph{noise} for convenience.

We consider two types of noise, both based on the fact that the rules operate on discretized features, e.g.~thresholded NetFractionRevolvingBurden in Section~\ref{sec:data:FICO}. Perturbations to these features may thus result in changes from one explanation or decision to a ``nearby'' one. The first type of noise models variability in the thresholds applied to discretize continuous features. This variability may arise, for example, from different loan officers' opinions about what constitutes a high versus a medium debt burden. Suppose that in the absence of noise, a continuous feature $X_j$ is converted into an ordinal feature $Z_j \in \{1,\dots,L\}$ by means of thresholds $t_0, t_1, \dots, t_L$ such that $Z_j = \ell$ if and only if $X_j \in (t_{\ell-1}, t_\ell]$ ($t_0$ and $t_L$ may be infinite). In the presence of noise, the mapping becomes random and is given by the conditional cumulative distribution function \begin{equation}\label{eqn:tau}
    \Pr(Z_j \leq \ell \:\vert\: X_j = x_j) = g\left( \frac{t_\ell - x_j}{\tau \sigma_{X_j}} \right), \quad \ell = 1,\dots,L,
\end{equation}
where $g$ is the logistic function and $\tau > 0$ is a parameter that scales the standard deviation $\sigma_{X_j}$ of $X_j$. If $x_j$ is much lower or much higher than the threshold $t_\ell$, then the probability is close to either $1$ or $0$. However, if $x_j$ lies in a transition zone around $t_\ell$, whose width is defined by $\tau \sigma_{X_j}$, then there is uncertainty as reflected in a probability between $0$ and $1$.

The second type of noise occurs after discretization and perturbs the value of an ordinal feature $Z_j$. This may model human errors in misreading $Z_j$ or more general variability that occurs in this discrete space (for example, one loan officer may have a rule concerning high-debt applicants while another includes both medium and high debt). It is assumed that perturbations to nearby values are most likely. Specifically, we generate a noisy feature $\tilde{Z}_j$ from $Z_j$ according to 
\begin{equation}\label{eqn:epsilon}
    \Pr(\tilde{Z}_j = k \:\vert\: Z_j = \ell) = \begin{cases}
    1 - \epsilon, & k = \ell\\
    \epsilon, & (k,\ell) = (1,0) \text{ or } (L-1,L)\\
    \epsilon / 2, & k = \ell \pm 1, \; 1 < \ell < L\\
    0 & \text{otherwise,}
    \end{cases}
\end{equation}
so that only mappings to adjacent levels are allowed, with total probability $\epsilon$. 

\begin{figure*}[ht]
    \centering
    \includegraphics[width=0.25\textwidth]{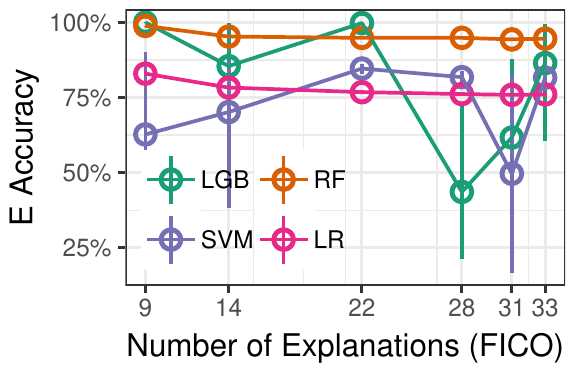}\hspace{2em}
    \includegraphics[width=0.25\textwidth]{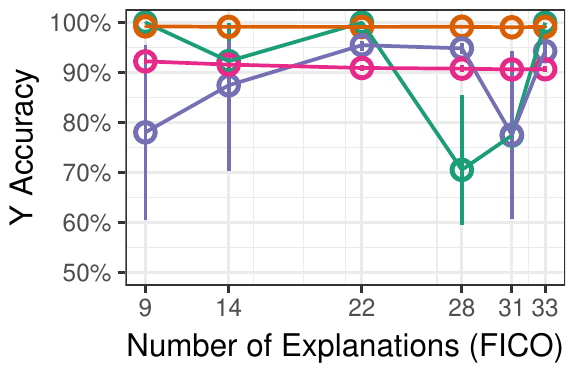}\hspace{2em}
    \includegraphics[width=0.25\textwidth]{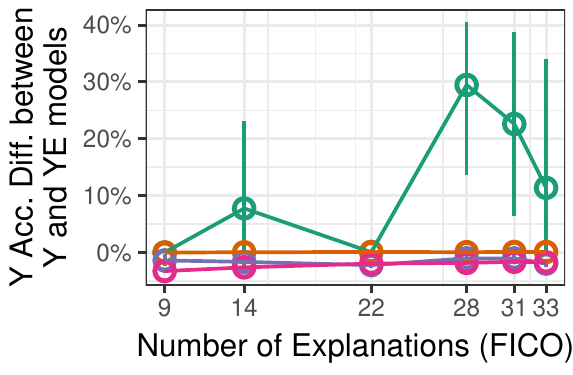}
    \caption{The effect of the number of possible explanations on TED-C's $E$ and $Y$ accuracy using four classification algorithms. The circles represent mean accuracy across five folds, and the error bars represent the 95\% confidence interval for the mean. The rightmost graph shows the $Y$ accuracy difference between a model that predicts only $Y$ and one that predicts $(Y, E)$ pairs.}
    \label{fig:fico_nexp_acc}
\end{figure*}

\begin{figure*}[th]
    \centering
    \includegraphics[width=0.25\textwidth]{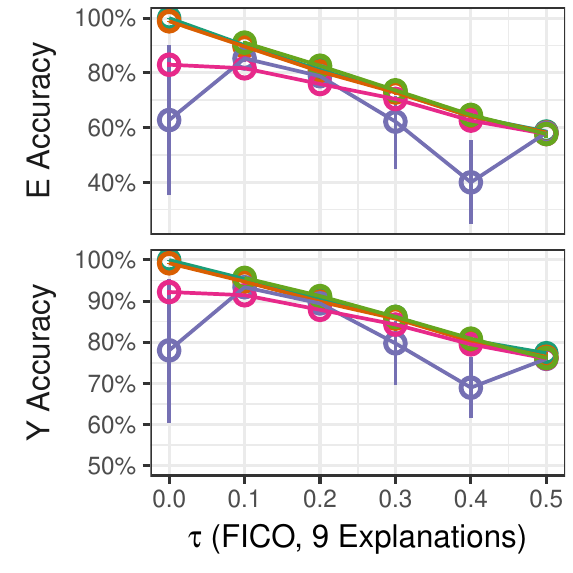}\hspace{2em}
    \includegraphics[width=0.25\textwidth]{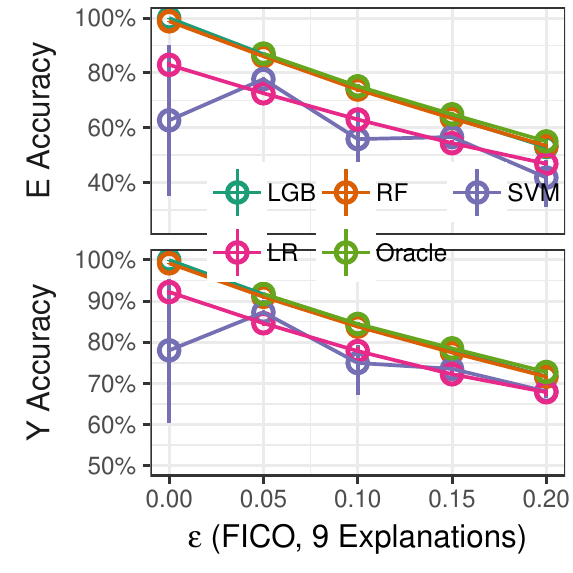}\hspace{2em}
    \includegraphics[width=0.25\textwidth]{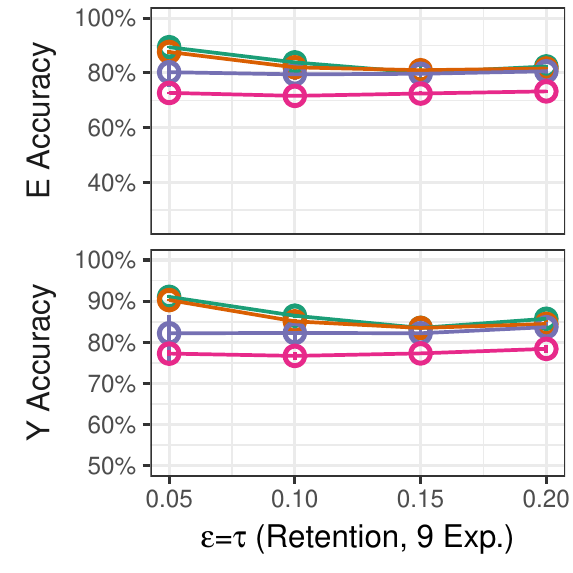}
    \caption{The effect of explanation noise parameters $\tau$ and $\epsilon$ on TED-C's $E$ and $Y$ accuracy.}
    \label{fig:fico_noise}
\end{figure*}

\section{Experiments}
\label{sec:expt}
This section evaluates the TED-C framework's performance under the impact of four variables: the number of explanation options, the amount of noise in the explanation ($E$) and decision ($Y$) labels, the availability of $E$ labels, and the specific algorithm used to classify the Cartesian product $(Y, E)$.

We report multiple measures of the performance of TED-C: $E$ accuracy, $Y$ accuracy, and the difference between $Y$ accuracy in predicting the $(Y, E)$ Cartesian product and in predicting $Y$ alone. The third measure answers our first research question, i.e., whether and how the TED-C framework affects a classifier's ability to predict decisions given the requirement to also predict explanations. Both accuracy (proportion of labels predicted correctly) and F1 scores were collected from the experiments. However, because we found that the two were highly correlated, we present only the accuracy results below.

We evaluated TED-C's performance using four classification algorithms: LightGBM (\textbf{LGB}, \citealp{LightGBM}), Random Forest (\textbf{RF}), \textbf{SVM}, and Logistic Regression (\textbf{LR}). The first two produce tree ensembles, with LGB being a state-of-the-art gradient boosting method, while the last two are linear classification algorithms, where SVM may use nonlinear kernels. We hypothesize that since our explanations were generated by rules and the first two algorithms are similar to rule-based algorithms, they would likely perform better than the linear classifiers within the TED-C framework. 

To attempt to maximize the 
performance of each algorithm, we conducted hyperparameter optimization using the Bayesian Optimization and Hyperband method \cite{falkner18a}. This method is efficient because it combines Bayesian optimization with bandit-based optimization. 
\ignore{
It also uses a successive halving strategy, which deploys models on a low-computation-requirement setting, such as with a small training set or fewer number of trees, to conduct random search, and then targets the promising hyperparameter regions with more computationally intensive settings.}  We did not employ its  successive halving feature because each run of these four algorithms is reasonably fast.
Table~\ref{tab:hyperparameters} (in Appendix~\ref{sec:expt:add}) lists, for each algorithm, the hyperparameters that were searched in the experiment. All other parameters were kept constant at their default values except 1) RF's number of estimators was set to 100 (default is 10), and 2) SVM and LR's maximum number of iterations was set to 1000 because little improvement in accuracy was observed beyond this point. 
However, this restriction was applied only in the parameter tuning phase and was removed during final model fitting for better accuracy.

For each experimental setting, a five-fold CV was conducted for each classification algorithm. Test results are averaged across these five folds and reported below. For each fold, we conducted hyperparameter search using the training data of that fold, and evaluated the resulting optimal model on the test data of that fold. The hyperparameter search was configured to optimize the classifier's $E$ accuracy since TED-C's primary purpose is to generate explanations. For SVM, we ran hyperparameter search for 50 iterations, whereas for the other three algorithms we ran the search for only 20 iterations since their hyperparameter spaces are not large. In each search iteration, an internal five-fold CV was conducted to generate an estimate of out-of-sample $E$ accuracy. The remaining parts of this section discuss the experimental results.

\begin{figure*}[h]
    \centering
    \includegraphics[width=0.25\textwidth]{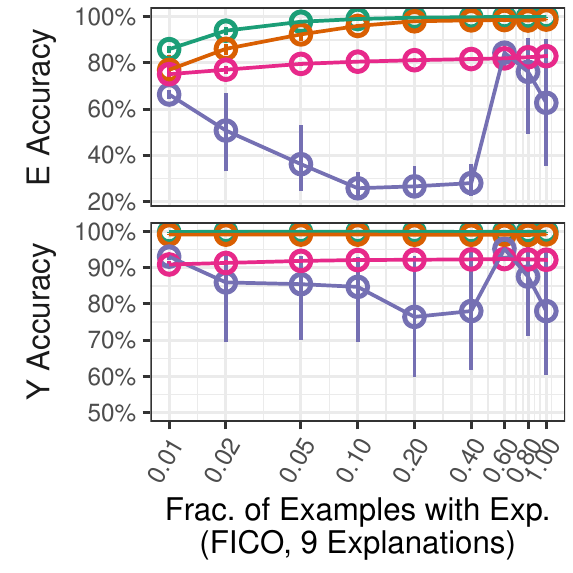}\hspace{2em}
    \includegraphics[width=0.25\textwidth]{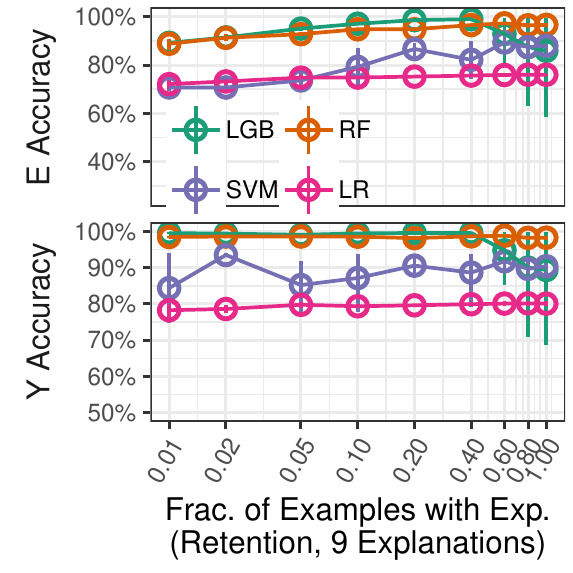}\hspace{2em}
    \includegraphics[width=0.25\textwidth]{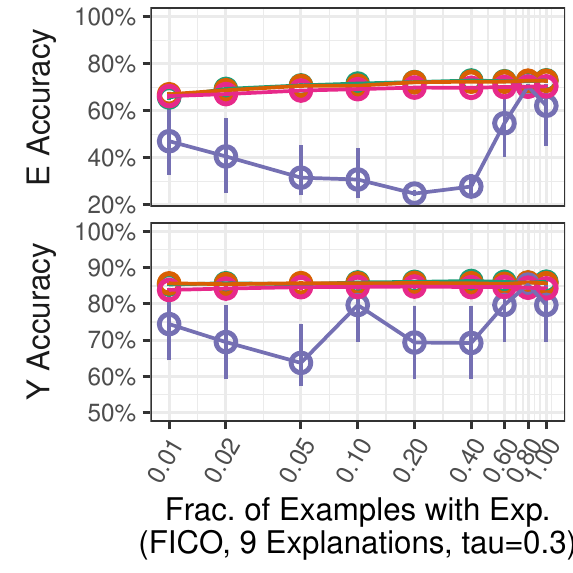}
    \caption{The effect of missing training explanations on TED-C's $E$ and $Y$ accuracy.}
    \label{fig:fico_semi}
\end{figure*}

\subsection{Increasing the Number of Explanations}
\label{sec:expt:number}
Under the TED-C approach, the set of possible explanations is enumerated and the size of the set translates directly into the number of classes that the classification algorithm must handle. It is natural to ask how performance may be affected as this number increases. We consider this question by increasing the number of explanations in the FICO and retention datasets beyond the base 9-explanation versions, as described in Section~\ref{sec:data:FICO} and \ref{sec:data:retention}.

Figure~\ref{fig:fico_nexp_acc} shows the impact of the number of possible explanations on both $E$ and $Y$ accuracy. The left two sets of graphs show that the accuracies of RF (brown) and LR (pink) stay relatively constant as the number of explanations increases. Impressively, RF was able to maintain almost 100\% accuracy on both $E$ and $Y$ across all settings. On the other hand, LGB and SVM were unstable when the number of explanations is large, as evidenced by their large error bars. In some cases, their average $E$ accuracy dropped to 50\%. This is somewhat surprising given LGB's state-of-the-art performance on many classification datasets,\footnote{See 
\url{https://github.com/microsoft/LightGBM/blob/master/docs/Experiments.rst#comparison-experiment}.} and SVM's usual outperformance over LR. Figure~\ref{fig:retention_nexp_acc} in Appendix~\ref{sec:expt:add} shows similar behavior on the retention dataset.

The rightmost graph in Figure~\ref{fig:fico_nexp_acc} shows how the TED-C framework may affect $Y$ accuracy under some settings. The graph shows the difference in $Y$ accuracy between a model that is built to  predict only $Y$ and a model under the TED-C framework that is built to predict $(Y, E)$ pairs. As can be seen, for all algorithms except LGB, the lines stay near 0\%, suggesting that TED-C's need to predict $(Y, E)$ pairs does not affect its ability to predict $Y$. In fact, for LR, the accuracy has improved by as much as 3.2\% with the introduction of $E$ labels, suggesting that the $E$ labels provided extra information for LR to predict $Y$. For LGB, however, a model that predicts only $Y$ can sometimes outperform its TED-C counterpart by 40\%. This again shows that LGB has a stability issue when there is a large number of classes.

\subsection{Increasing Variability in Explanations}
\label{sec:expt:noise}

Figure~\ref{fig:fico_noise} shows the effects on TED-C's accuracy of adding the two types of noise described in Section~\ref{sec:data:noise} and parametrized by $\tau$ and $\epsilon$ respectively. These are added separately to the FICO dataset with 9 explanations and jointly to the retention dataset with 9 explanations. In addition to curves for the four classification algorithms, we also plot the performance of an ``oracle'' that knows the noiseless rules in Sections~\ref{sec:data:FICO} and \ref{sec:data:retention} for producing explanations and decisions, i.e., without having to learn them from data. The oracle is thus suggestive of the highest accuracy possible for a given noise level. Both RF and LGB perform very close to the oracle, with LGB not exhibiting the instability seen in Figure~\ref{fig:fico_nexp_acc}. LR is worse, as expected, due to its linearity but also degrades gracefully, and even closes the gap with respect to the oracle in the FICO $\tau$ plot. The performance of SVM remains uneven.

\subsection{Reducing Required Explanations}
\label{sec:expt:partial}

\ignore{
Another potential concern with the TED-C framework is the additional burden placed on human experts to provide explanation labels as well as decision labels. In particular, we assume that labelers are accustomed to providing decisions but may be less able to give explanations for all training examples.}

This subsection explores scenarios where only a subset of the training data has explanations.
We simulate this with the FICO and retention datasets by removing at random a fraction of the $E$ labels from the training sets.  We use the algorithm described in Section~\ref{sec:ted-partial} to impute missing $E$ labels.
To simplify the experiment, the same classification algorithm (e.g.~LGB, SVM) is used for imputation as for TED-C. For imputation, two separate models are trained for $Y=0$ and $Y=1$ (with $X$ as input), instead of appending $Y$ as an input alongside $X$. This takes advantage of the fact that $Y$ is binary and restricts the set of possible $E$'s to those associated with either the positive or negative class.

Figure~\ref{fig:fico_semi} shows the accuracies obtained as a function of the fraction of training $E$ labels that are given. Remarkably, accuracies decrease only slightly as the fraction decreases to $0.1$ ($10\%$ of training data have explanations) and moderately even at $0.01$. The exception is SVM, which is again unstable. The rightmost plot shows that accuracies are maintained even in the presence of noise ($\tau = 0.3$). A possible explanation is that the explanation concepts in these datasets are logical and do not require many examples to learn.

\section{Discussion and Open Problems}
\label{sec:discuss}

The experimental results in Section~\ref{sec:expt} have shown that, insofar as human explanations and decisions follow logical rules, the TED-C approach can reproduce them with high accuracy. This is true even as the number of possible explanations increases to more than $30$ or as the fraction of training data with explanation labels decreases below $10\%$. In the presence of noise, accuracies degrade gradually and nearly match those of an oracle that knows the noise-free generating process. Of the four classification algorithms compared, Random Forest stands out for achieving high accuracies and stability across the tested conditions, while the others suffer from lower accuracy (logistic regression), instability (LightGBM), or both (SVM). With the exception of LightGBM, accuracy in predicting class labels $Y$ does not decrease with the addition of explanation labels $E$ and may in fact improve.

We have considered the case where class and explanation labels are produced at the same time. As mentioned in Section~\ref{sec:TED}, in the case where class labels $Y$ have already been assigned by another entity or process, it can be challenging to give explanations $E$ that are consistent with the class labels. Assuming that explanations can be added after the fact, whether consistent or not, TED-C can be applied in exactly the same way to these $(Y,E)$ labels to make predictions. The open problem to be addressed is how to handle instances for which the \emph{predicted} $Y$ is inconsistent with the predicted $E$.  One option is to use the partially supervised learning described in this work to impute explanations from the assigned labels.

Another area of future work is to pursue use cases where the training explanations are produced by humans rather than automatically constructed from rules. One possibility is to crowd-source explanations, although this falls short of the ideal of having explanations from true experts.  Another scenario is to have a human
produce labels and explanations based on rules written by experts.  This can be useful in dealing with ambiguous or ill-defined rules because the human can apply their intuition to resolve the confusion.  However, humans are  also prone to mistakes, i.e., misapplying a rule, which could lead to lower quality training data.  Either way, this will be an interesting scenario to consider in the future.

\ignore{The fraction of inconsistent predictions may be much lower than the fraction of inconsistent training labels. For example, using FICO data where $E$ is generated by the rules in Section~\ref{sec:data:FICO} but $Y$ is the original repayment label, the inconsistency in training labels is $29\%$ (since the rules are $71\%$ accurate w.r.t.~repayment) but is only ?? for the predicted labels.}


\bibliographystyle{aaai}
\bibliography{ted,ExAbsent}

\clearpage
\appendix
\appendixpage

\begin{figure*}[!tb!]
    \centering
    \includegraphics[width=0.33\textwidth]{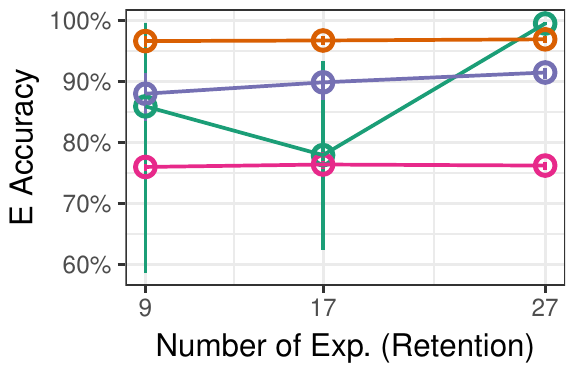}
    \includegraphics[width=0.33\textwidth]{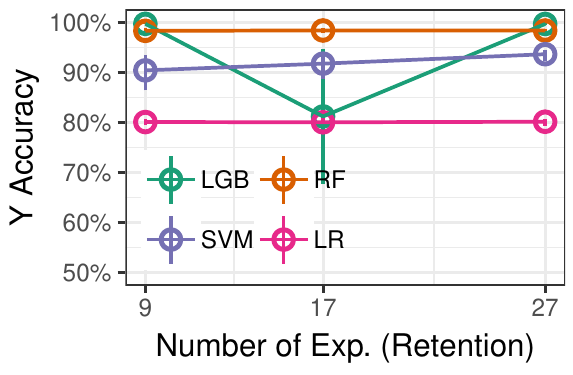}
    \includegraphics[width=0.33\textwidth]{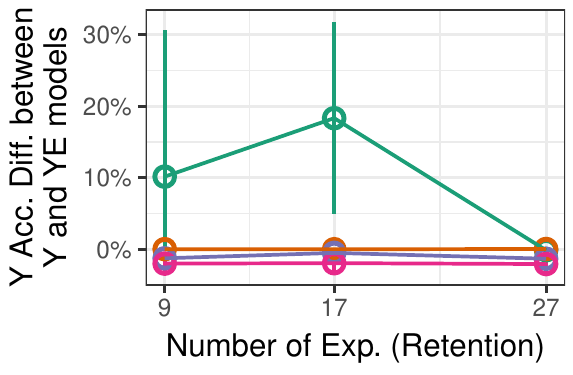}
    \caption{The effect of the number of explanations on TED-C's $E$ and $Y$ accuracy for the retention dataset. The rightmost graph shows the $Y$ accuracy difference between a model that predicts only $Y$ and a model that predicts $(Y, E)$ pairs.}
    \label{fig:retention_nexp_acc}
\end{figure*}

\section{Additional FICO Dataset Details}
\label{sec:data:FICOadd}

The original FICO challenge dataset contains three kinds of special values: $-9$ (no record), $-8$ (no usable/valid trades or inquiries), and $-7$ (condition not met). First we removed $588$ rows where all features have value $-9$ and are thus not usable, leaving $9871$ rows. After removing the ExternalRiskEstimate feature as mentioned in Section~\ref{sec:data:FICO}, no $-9$ values remain in the data. Next, only two features have $-7$ values: MSinceMostRecentDelq and MSinceMostRecentInqexcl7days. By analyzing the rates of repayment as a function of these two features, we concluded that MSinceMostRecentDelq $= -7$ most likely means that the applicant did not have a delinquency in the past 7 years (and hence ``condition was not met''), while MSinceMostRecentInqexcl7days $= -7$ means that the applicant in fact had a credit inquiry in the last 7 days, which was excluded. Hence we replaced MSinceMostRecentDelq $= -7$ with $84$ months (7 years) and MSinceMostRecentInqexcl7days $= -7$ with $0$ months. The remaining special values are all $-8$, which we replaced with the generic \texttt{np.nan} null value in NumPy. Some classification algorithms (e.g.~LGB) can handle null values while for LR and SVM, null values were further imputed using the median value of the feature. Lastly, we were able to infer that all cases of MaxDelq2PublicRecLast12M $> 7$ (``other values'') should actually be equal to $7$ (current and never delinquent in last 12 months) based on the corresponding value of MaxDelqEver, which implied that the applicant was never delinquent.

Below we show the exact rule set learned by BRCG with regularization parameters $\lambda_0 = \lambda_1 = 10^{-5}$, which differs from the rule set in Section~\ref{sec:data:FICO} only in the threshold values.
\begin{enumerate}
    \item NetFractionRevolvingBurden $\leq 63$ AND\\ PercentTradesNeverDelq $> 86$ AND\\ AverageMInFile $> 52$ AND\\ MaxDelq2PublicRecLast12M $> 5$; OR
    \item NetFractionRevolvingBurden $\leq 39$ AND\\ MSinceMostRecentInqexcl7days $> 24$.
\end{enumerate}

\section{Additional Experimental Details and Results}
\label{sec:expt:add}

Table~\ref{tab:hyperparameters} lists the hyperparameters that were optimized for each classification algorithm. Figure~\ref{fig:retention_nexp_acc} shows the effect of the number of explanation choices on TED-C's $E$ and $Y$ accuracy for the retention dataset.

\begin{table}[h!]
\begin{small}
    \caption{Hyperparameter search space definitions.}
    \centering
    \begin{tabular}{rrl}
         & Hyperparameter & Search space \\
        \toprule
        LGB & min\_child\_samples & $1$--$50$\\
        \midrule
        RF & min\_samples\_leaf & $1$--$50$\\
        \midrule
        SVM & kernel & linear, polynomial\\
        & & RBF, sigmoid\\
        & C & $10^{-3}$ to $10$, log scale \\
        & degree & $2$--$4$\\
        & gamma & $10^{-5}$ to $1$, log scale\\
        \midrule
        LR & penalty & $\ell_1$, $\ell_2$\\
         & C & $10^{-3}$ to $10$, log scale\\
    \end{tabular}
    \label{tab:hyperparameters}
\end{small}
\end{table}

\end{document}